\documentclass{article}

\usepackage[final,nonatbib]{neurips_2022}

\usepackage[utf8]{inputenc} 
\usepackage[T1]{fontenc}    
\usepackage[hidelinks]{hyperref}       
\usepackage{url}            
\usepackage{booktabs}       
\usepackage{amsfonts}       
\usepackage{nicefrac}       
\usepackage{microtype}      
\usepackage{xcolor}         

\usepackage{graphicx} 
\usepackage{caption,subcaption}
\usepackage{float}
\usepackage[free-standing-units=true,per-mode=symbol]{siunitx}

\usepackage{algorithm}
\usepackage{algorithmic}
\usepackage{amsmath,amssymb}
\usepackage{cleveref}

\usepackage{svg}
\usepackage[style=ieee,natbib=true,doi=false,isbn=false,url=false,dashed=false,mincitenames=1,maxcitenames=1]{biblatex}
\bibliography{bib.bib}
\AtNextBibliography{\small}

\usepackage{tikz}
\usepackage{pgfplots}

\definecolor{pastelMagenta}{HTML}{FF48CF} 
\definecolor{pastelPurple}{HTML}{8770FE} 
\definecolor{pastelBlue}{RGB}{0,114,178} 
\definecolor{pastelSkyBlue}{RGB}{86,180,233} 
\definecolor{pastelSeaGreen}{RGB}{86,180,233} 
\definecolor{pastelGreen}{RGB}{0,158,115} 
\definecolor{pastelOrange}{RGB}{230,159,0} 
\definecolor{pastelRed}{HTML}{F5615C} 
\definecolor{darkColor}{HTML}{300A24} 

\pgfplotsset{compat=newest}
\pgfplotsset{every axis legend/.append style={legend cell align=left}}

\title{Graph Q-Learning\\ for Combinatorial Optimization}

\author{%
  Victoria M.~Dax \\
  Stanford University\\
  \texttt{vmdax@stanford.edu} \\
   \And
   Jiachen Li \\
   Stanford University\\
   \texttt{jiachen\_li@stanford.edu} \\
   \And
   Kevin Leahy\\
   MIT Lincoln Laboratory\\
   \texttt{kevin.leahy@ll.mit.edu}
   \And 
   Mykel J. Kochenderfer \\
   Stanford University\\
   \texttt{mykel@stanford.edu} \\
}

\begin{document}

\maketitle

\begin{abstract}
Graph-structured data is ubiquitous throughout natural and social sciences, and Graph Neural Networks (GNNs) have recently been shown to be effective at solving prediction and inference problems on graph data. In this paper, we propose and demonstrate that GNNs can be applied to solve Combinatorial Optimization (CO) problems. CO concerns optimizing a function over a discrete solution space that is often intractably large. To learn to solve CO problems, we formulate the optimization process as a sequential decision making problem, where the return is related to how close the candidate solution is to optimality. We use a GNN to learn a policy to iteratively build increasingly promising candidate solutions. We present preliminary evidence that GNNs trained through Q-Learning can solve CO problems with performance approaching state-of-the-art heuristic-based solvers, using only a fraction of the parameters and training time.
\end{abstract}

\section{Introduction}\label{intro}

Many important real-world problems, from social networks to chemical systems, are naturally represented as graphs. Over the past five years, graph neural networks (GNNs) have emerged as a powerful approach for machine learning on such structural data \cite{gori2005new, scarselli2008graph} by combining the representation learning capabilities of deep networks with bespoke adaptation based on graph properties. These approaches have been effective in node classification \cite{kipf2017semisupervised}, relation prediction \cite{ying2018graph}, and graph classification \cite{morris2019weisfeiler}. More recently, the community has been exploring other applications \cite{zhao2021adaptive} for GNNs and analyzing their theoretical limitations and mathematical properties \cite{fereydounian2022exact}. 

Many of these successes have focused on prediction or inference problems. In this work, we demonstrate that GNNs can be applied to solve a different class of problems. Combinatorial Optimization (CO) problems are often encountered across diverse fields and are difficult to solve exactly. CO problems require optimizing a function over a combinatorial space, possibly subject to constraints; examples include finding the minimum spanning tree (MST) in a graph or determining if there exists a variable assignment that satisfies a given Boolean formula ($k$-SAT). The ubiquity of CO across domains and applications led to the development of heuristic-based solvers in the early days of computer engineering that remain state-of-the-art today \cite{MILP}.

Motivated by deep learning's success at learning representations that outperform hand-engineered features, we explore whether GNNs can learn to outperform heuristic-based CO solvers when trained via reinforcement learning (RL). Our specific contributions are:
\begin{enumerate}
    \item Representing instances of combinatorial optimization problems as graphs and formulating solving for an optimized solution as a Markov decision process (MDP). To the best of our knowledge, this work is the first to solve FJSP using a deep learning-based method.
    \item  Showing how GNNs can be used to solve CO problems and that our formulation generalizes to a form of meta-learning.
    \item Demonstrating empirically that we meet the performance of other algorithms and baseline heuristics with a fraction of the parameters and training time. 
\end{enumerate}

\section{Literature Review}

The current research landscape in the field of graph based learning consist of theoretical improvements to the architecture \cite{xu2018representation, brody2021attentive}, explorations of limitations \cite{alon2020bottleneck}, and applications to a variety of domains through supervised learning.

Graph neural networks have only very recently transitioned into usage in the context of RL. However rather than end-to-end, GNNs are employed in a modularized fashion: \citet{deep_implicit_coordination_graph} use a GNN for inferring a graph given sequential data streams. \citet{reward_redistribution} employ a graph attention network (GAT) in a multi-agent RL context, but for optimal reward balancing to speed up the learning process. \citet{game_abstraction} is using a GNN for game abstraction to simplify complex multiplayer games. \citet{klogit} leverage the node update to mimic k-logit in a two-player zero-sum games.

Very few papers explore GNNs in the context of CO \cite{mazyavkina2021reinforcement}. These works focus on solving the travelling salesperson problem only, tailoring their implementation to fit this task specifically, and use GNNs as a search heuristics to prune the solution space rather than training a solver directly.  

\section{Preliminaries}

\textbf{Combinatorial optimization} (CO) refers to optimizing an objective function whose domain is a discrete but combinatorially large configuration space, making the space of possible solutions typically too large to search exhaustively. Examples of well-known combinatorial optimization problems include the Travelling Salesman Problem (TSP), Minimum Spanning Tree (MST), and Boolean Satisfiability (SAT). While some instances of CO problems can be solved exactly through Branch-and-Bound, many are NP-Hard. We generally must resort to specialized heuristics that rule out large parts of the search space or approximation algorithms. Formally, a combinatorial optimization problem $A$ is defined by the tuple $(\mathcal{I},f,m,g)$, where $\mathcal{I}$ is a set of instances, $x \in \mathcal{I}$ is an instance, $f(x)$ is the finite set of feasible solutions $y$, $m(x,y)$ denotes the measure of $y = f(x)$, and $g$ is the goal function, i.e. usually $\max$ or $\min$. The goal is then to find for some instance $x$ an optimal solution, that is, a feasible solution $y$ with
\begin{equation}
m(x,y)=g\{m(x,y')\mid y'\in f(x)\}.
\end{equation}

In \textbf{graph-based learning}, a GNN layer can be viewed as a message-passing step \cite{gilmer2017neural}, where each node updates its state by aggregating messages flowing from its direct neighbors.
A graph is a tuple of nodes and edges $\mathcal{G} = (\mathcal{V}, \mathcal{E})$. The one-hop neighborhood of node $u$ is $\mathcal{N}_u = \{v \in V \mid (v, u) \in \mathcal{E}\}$. A node feature matrix $X \in R^{|V|\times k}$ gives the $k$ features of node $u$ as $x_u$; we omit edge- and graph-level features for clarity. A (message passing) GNN over this graph is then executed as: 
\begin{equation}
    \textbf{h}_u = \psi_{\textsc{AGG}}\big(\{\phi(x_u, x_v) | v \in \mathcal{N}_u\}\big) 
\end{equation}

where $\psi \colon R^k \times R^k \rightarrow R^k$ is a message function, $\phi \colon R^k \rightarrow R^k$ is a readout function, and $\textsc{AGG}$ is a permutation-invariant aggregation function (such as $\Sigma$ or $\max$). Both $\phi$ and $\psi$ can be realised as MLPs, but many special cases exist, such as attentional GNNs \cite{velivckovic2017graph}.

\textbf{Reinforcement Learning} (RL) considers an agent learning how to select actions in an environment to maximize their long-term cumulative rewards, i.e., the return, in a sequential decision-making process \cite{kochenderfer2022algorithms}. The environment is modeled as a Markov Decision Process (MDP), defined by the tuple ($\mathcal{S}$, $\mathcal{A}$, ${p}$, ${r}$, $\gamma$) with $\mathcal{S} = \{s\}$ the set of states, $\mathcal{A} = \{a\}$ the set of actions, ${p}(s' \mid s, a)$ the state transition distribution, ${r} \colon \mathcal{A} \times \mathcal{S} \rightarrow \mathcal{R}$ a bounded reward function, and $\gamma$ the discount factor. RL aims to find a policy $\pi \colon \mathcal{S} \rightarrow \mathcal{A}$ that maps a state $s$ to an optimal action $a$. Optimality is defined as maximizing the expected return.

Model-free RL refers to learning a policy or Q-function, though a form of trial-and-error, without explicitly modeling the transition probability distribution ${p}$ or the reward function $r$. Deep Q-learning \cite{van2016deep} is a simple algorithm but can suffer from overestimation bias and catastrophic forgetting.
Several other methods have been introduced to reduce overestimation bias. Weighted double Q-learning \cite{zhang2017weighted}, for example, uses a weighted combination of the double Q-learning estimate, which may lead to underestimation bias.\\

\section{Method} \label{sec:method}

We frame combinatorial optimization problems as sequential decision making processes, where the return is related to how close a candidate solution is to optimal. We use a GNN to learn a policy that sequentially builds better and better candidate solutions. We aim to solve general classes of CO problems, but in this work we focus on the flexible Job Shop Scheduling problem (FJSP).

\subsection{Problem Definition}

The classical job-shop scheduling problem (JSSP) \cite{applegate1991computational} defines $n$ jobs that consist of sets of operations $o_{i}$ of varying processing time each, which need to be processed in a specific order as specified by a set of precedence constraints. Each operation is assigned to one of $m$ machines and the goal is to find a processing schedule that minimizes the makespan, i.e., the total length of the schedule from the start of the first job to the end of the last. The FJSP \cite{chaudhry2016research} is a generalization of the classical JSSP that allows processing operations on one machine out of a set of alternative machines. The FJSP is an NP-hard problem consisting of two sub-problems, i.e., the assignment and the scheduling problems. The scheduling problem by itself reduces to the classical JSSP. Viable solutions are evaluated in terms of the makespan, which is defined as the time difference between start and finish of a sequence of jobs or tasks.

\begin{figure}[h]
    \centering
    \includegraphics[width=0.85\linewidth]{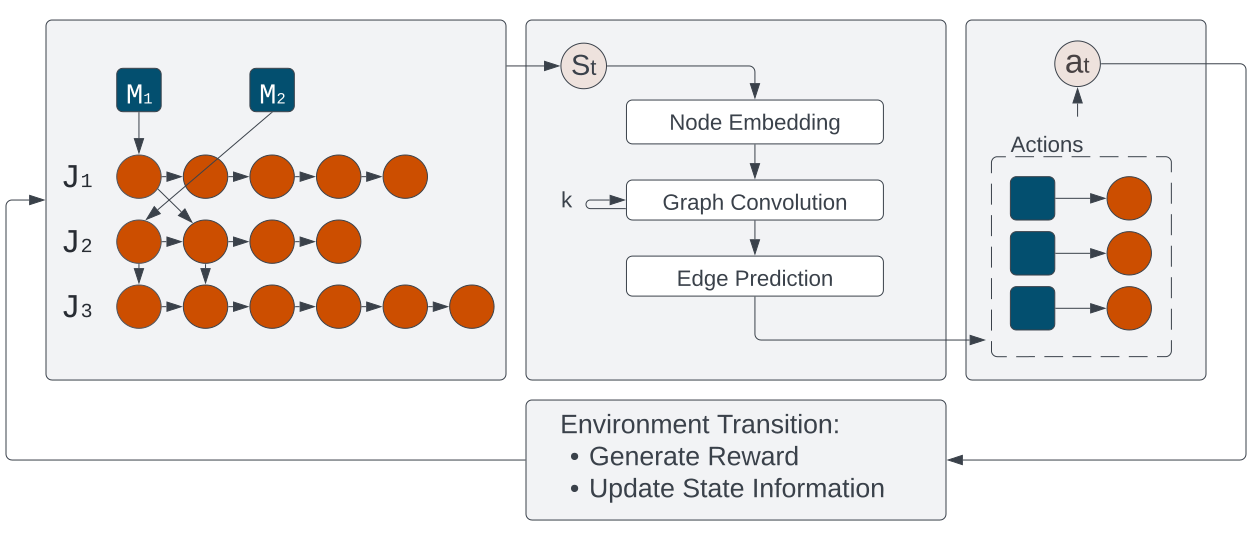}
    \caption{Training cycle.}
    \label{fig:training_cycle}
\end{figure}

In this paper, proposing a solution to an instance of the FJSP is treated as a sequential decision-making process, which iteratively takes a scheduling action to assign an operation to a compatible machine at each state until all operations are scheduled. The proposed workflow is shown in \Cref{fig:training_cycle}.

\subsection{Combinatorial Optimization as MDPs} \label{ssec:mdp}

By reformulating FJSP as a sequential decision making problem, more specifically a Markov Decision Process (MDP), we can use traditional RL methods to find viable solutions to a given FJSP. The initial state $s_0$ defines the problem setting, i.e., it defines how many machines are available, how many operations will need to be processed, and in what sequence. At each time step $t$, an action $a_t$ specifies the identity of an unassigned operation to be added to the end of the queue of a machine. Consequently, a state $s_t$ defines the resource allocation and processing sequence. A sequence ends when all operations of all jobs have been assigned to a machine, which will be the proposed solution to the problem instance. If the agent assigns operations with upstream requirements first, these operations can not be executed and render a machine idle until its requirements are met. Gridlock defines the state where all machines are idle by assignment.

During a rollout, previously assigned jobs can start processing and the agent will receive positive unit rewards for completed operations, which are then removed from the queue. To encourage solving for a shorter makespan, a negative reward of $-0.1$ is accrued at each time step. 

An action-value function $Q(s_t, a_t)$ maps the expected return of taking action $a_t$ in state $s_t$. This next section defines the architecture of a heterogeneous graph neural network that we subsequently train to learn $Q$.

\subsection{Heterogeneous Graph Neural Networks} \label{ssec:gnn}

We employ a disjunctive graph $\mathcal{G} = \{\mathcal{O}, \mathcal{M}, \mathcal{E}_j, \mathcal{E}_q, \mathcal{C}\}$ to model the current state of a FJSP. Here, $\mathcal{O}$ is the set of operations independent of the job to which they belong, $\mathcal{M}$ is the set of available machines, $\mathcal{E}_j$ and $\mathcal{E}_q$ are sets of directed edges that denote the sequence of operations within each job and within each queue respectively, and $\mathcal{C}$ represents the set of conjunctive, undirected edges that assign operations to machines.

\begin{figure}[h]
  \centering
  \begin{subfigure}{0.45\linewidth}
    \centering
    \includegraphics[width=0.9\linewidth]{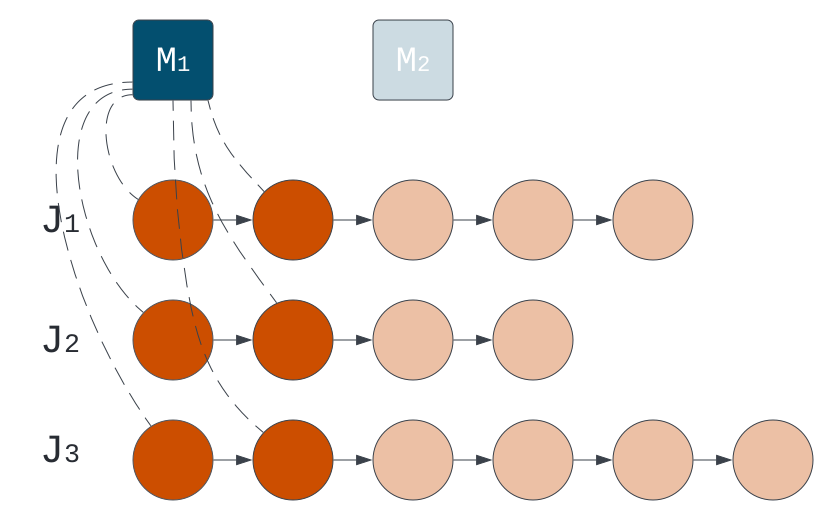}
    \caption{Initial state representation. \label{subfig:s0}}
  \end{subfigure}
  \begin{subfigure}{0.45\linewidth}
    \centering
    \includegraphics[width=0.9\linewidth]{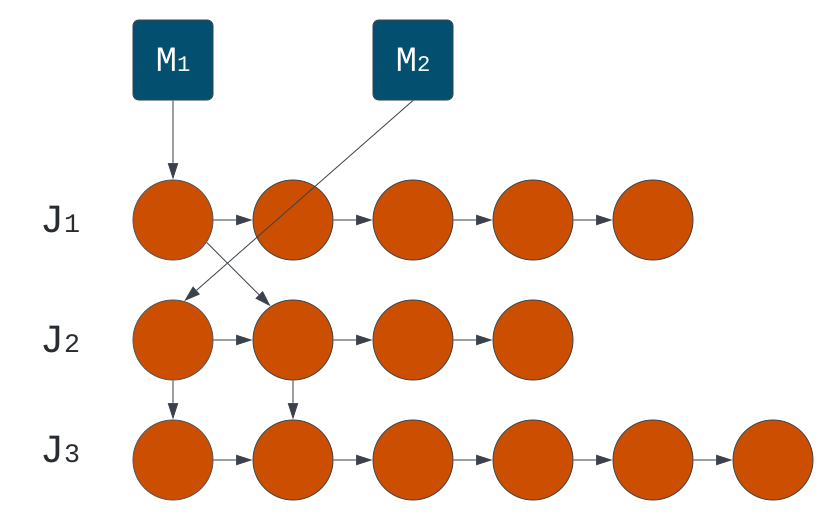}
    \caption{\label{subfig:s5} State at $t=5$}
  \end{subfigure}
  \caption{Graphical representation of a sample FJSP instance at $t=0$ (a) and after 5 actions have been taken, $s_{t=5}$ (b).}
    \label{fig:state_rep}
\end{figure}

Edges in $\mathcal{C}$, represented by dashed edges in \Cref{subfig:s0}, specify operation-machine compatibility and their assigned weights define a speed up or slow down of the default processing time. The feature space of operations is composed of i) the time required to finish the operation, ii) its completion percentage, iii) the number of downstream dependencies, iv) a one-hot encoding of the current state of this operation, i.e., whether it is scheduled, being processed, or completed, and v) the remaining time. The machine features comprise the number of queued operations and their minimal expected run time. 

To encode this state representation into a meaningful action-value function $Q$, we use a heterogeneous graph neural network composed of graph convolution layers \cite{kipf2017semisupervised, sageconv}. Our network architecture consists of two fully connected layers to embed both node types into the same dimensionality, followed by a set of convolutional layers, each for processing a different edge-type without weight sharing. The resulting intermediate node embeddings are summed. By looping over this same heterogeneous layer $k$ times, each node embedding considers the state information of nodes within a $k$ step radius. A dot-product readout layer is then used for edge-prediction over $\mathcal{C}$. The edge with the highest score defines the new operation-to-machine assignment, shown by solid blue arrows in \Cref{subfig:s5}.

\section{Experimental Results}

\subsection{Implementation details}

Random FJSP samples are created by initializing $m$ machines and $n\cdot \Tilde{n}_o$ operation nodes, and then queuing them randomly into $n$ jobs. Here, $\Tilde{n}_o$ refers to the average number of operations per job. To create $\mathcal{C}$, we fully connect each machine to each operation, to indicate possible assignments, and then randomly drop a fraction $p$. Their weights indicate the relative speedup or slowdown for endpoint operations. Each operation is assigned a baseline runtime, which results in the actual runtime of operation $o_i$ on machine $w_j$ when adjusted by the connected edge weights.

For the following experiments, we set the number of HGNN iterations to $k=2$ and dimensions of machine and operation embeddings to $16$. In each epoch, we sample $128$ trajectories, which are stored in a replay buffer of size $5000$, and run $64$ training iterations with a batch size of $32$ state transitions. The discount factor is set to $\gamma = 1.0$ and the explorations constant to $\epsilon = 0.1$.
The network is updated using the Adam optimizer, with a learning rate of \num{8e-5}.

\subsection{Baselines}

To evaluate our method, we benchmark against simulated annealing, a probabilistic approximation method, and a state-of-the-art meta-heuristic introduced in \citet{fexlible_fjsp_metaheuristic}. To the best of our knowledge, this work is the first to solve FJSP using a deep learning-based method. We also benchmark our scheduling performance against recent deep learning methods designed for the simpler scheduling problem \cite{schedulenet}, JSSP.

To compare performance across methods, we evaluate the optimality gap
\begin{equation}
    \epsilon = \frac{C_{\min}}{C^*} - 1
\end{equation}

where $C_{\min}$ is the makespan of a candidate solution and $C^*$ the optimal makespan. This metric can also be referred to as relative error. Throughout these experiments, we use Google's OR-tools solver \cite{google_ortools} to solve for $C^*$. 

\subsection{Results}
\begin{figure}[H]
    \hspace*{-0.5cm}
    \begin{subfigure}{0.44\linewidth}
    \centering
        \begin{tikzpicture}[font=\footnotesize]
        \begin{axis}[width=7cm, height=5cm, 
                    ylabel={Loss},
                    xtick pos=bottom, ytick pos=left]
        \addplot[pastelBlue] table[col sep=comma,
        x=epoch,y=loss] {data/jobshop_qf_j25-w15_4x4_multilayer-from_script_v3-nan.csv};
        \end{axis}
        \end{tikzpicture}
    \end{subfigure}%
    \hfill
    \begin{subfigure}{0.44\linewidth}
    \centering
        \begin{tikzpicture}[font=\footnotesize]
        \begin{axis}[width=7cm, height=5cm,
                    ylabel={Reward},
                    xtick pos=bottom, ytick pos=left,
                    legend pos=south east,
                    legend style={nodes={scale=0.75, transform shape}}]
        \addplot[pastelBlue] table[col sep=comma,
        x=epoch,y=r_train] {data/jobshop_qf_j25-w15_4x4_multilayer-from_script_v3-nan.csv};
        \addplot[pastelRed] table[col sep=comma,
        x=epoch,y=r_eval] {data/jobshop_qf_j25-w15_4x4_multilayer-from_script_v3-nan.csv};
        \legend{Train, Eval}
        \end{axis}
        \end{tikzpicture}
    \end{subfigure}
    \vskip \baselineskip
    \begin{subfigure}{0.44\linewidth}
    \centering
        \begin{tikzpicture}[font=\footnotesize]
        \begin{axis}[width=7cm, height=5cm, ymin=-0.05, 
                    xlabel={Epoch}, ylabel={Success Rate},
                    xtick pos=bottom, ytick pos=left]
        \addplot[pastelBlue] table[col sep=comma,
        x=epoch,y=success_rates] {data/jobshop_qf_j25-w15_4x4_multilayer-from_script_v3-nan.csv};
        \end{axis}
        \end{tikzpicture}
    \end{subfigure}
    \hfill 
    \begin{subfigure}{0.44\linewidth}
    \centering
        \begin{tikzpicture}[font=\footnotesize]
        \begin{axis}[width=7cm, height=5cm, ymin=-0.05,
                    xlabel={Epoch}, ylabel={Relative Error},
                    xtick pos=bottom, ytick pos=left,
                    unbounded coords=jump]
        \addplot[pastelBlue] table[col sep=comma,
        x=epoch,y=relative_err] {data/jobshop_qf_j25-w15_4x4_multilayer-from_script_v3-nan.csv};
        \end{axis}
        \end{tikzpicture}
    \end{subfigure}
    \caption{Training performance summary ($25\times15$).}
    \label{fig:training_perforance}
\end{figure}
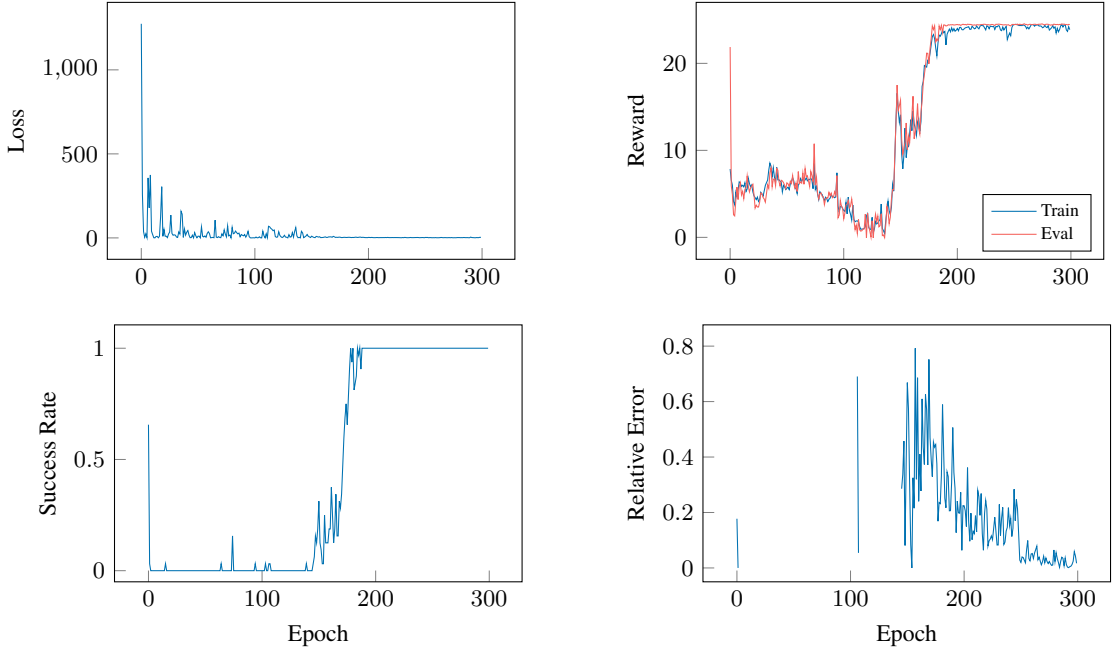

\Cref{fig:training_perforance} shows the learning curves for our Q-learner trained on sample problems of size $25\times15$. Around epoch 150, we find sudden jumps in the success rate and training rewards. These jumps happen when the learnt solver transitions from gridlocking itself to producing feasible solutions for the specified problem instances. The optimality gap can only be evaluated for feasible solutions, therefore the relative error curve starts around that same epoch and then decreases sharply as the solver learns to improve on its general strategy. 

After the Q-learner has been trained, it can be used to solve problems of any size. As such, this formulation can be interpreted as a type of meta-learning, enabled by using a graphical representation of the problem space that is not limited to a fixed problem size. We ran the same solver on multiple size problems and evaluated 128 samples in each case. \Cref{tab:opt_gaps} summarizes our results. We find that the meta-heuristic is a strong baseline that solves optimally until the largest size problem $100\times20$, while FIFO is not performing well from the beginning. The results reported for DQL are from the same network, but evaluated on different sample sizes. $100\times20$ is an exception where the relative error was $29\%$ but after 100 epochs of fine-tuning on larger problems, the error was reduced to $6.2\%$.

For further reference, we also report optimality gaps for ScheduleNet \cite{schedulenet}, which is a similar deep RL approach but for the classic scheduling problem JSSP. Our network architecture only defines 960 independent weights, while ScheduleNet defines 6022 for the actor alone. Because ScheduleNet adopts PPO, it further requires a trained critic. Therefore, we find that we maintain equal or better performance than an equivalent deep RL approach performs on a simpler problem, while using less than a sixth of its parameters.

\begin{table}[h]
        \caption{Optimality gaps for different FJSP sizes. \label{tab:opt_gaps}}
    \centering
    \begin{tabular}{@{}lrrrrrr@{}} \toprule
         & $15\times 15$ & $15\times 25$ & $30\times 20$ & $50\times 15$ & $50\times 20$ & $100\times 20$ \\ \midrule
        FIFO & -- & 0.7647 & 0.69 & 0.857 & -- & 1.235\\
        Meta-Heuristic & -- & 0.0 & 0.0 & 0.0 & 0.0 & 0.022\\
        DQL & 0.01 & 0.011 & 0.0 & 0.052 & 0.04 & 0.062\\
        ScheduleNet & 0.153 & 0.194 & 0.187 & 0.138 & 0.135 & 0.066  \\
        \bottomrule
    \end{tabular}
\end{table}

\begin{figure}[h]
    \centering
    \hspace*{-0.5cm}
    \begin{tikzpicture}[font=\footnotesize]
    \begin{axis}[legend pos=south east, 
                xlabel={Problem Size},
                ylabel={Runtimes [\si{\second}]},
                width=8.5cm, height=5.5cm,
                xticklabels={, 15×15, 15x25, 30×20, 50×15, 50×20, 100×20},
                legend pos=north west,
                legend style={nodes={scale=0.75, transform shape}},
                xtick pos=bottom, ytick pos=left]
    \addplot[pastelRed, draw] coordinates {(0, 0.102) (1, 0.171) (2, 0.205) (3, 0.347) (4, 0.347) (5, 0.622)};
    \addplot[pastelBlue, draw] coordinates {(0, 1.012) (1, 1.335) (2, 2.085) (3, 2.74) (4, 5.25) (5, 7.194)};
    \addplot[pastelGreen, draw] coordinates {(0, 0.0012) (1, 0.0023) (2, 0.0035) (3, 0.0066) (5, 0.027)};
    \legend{DQL, Meta-Heuristic, FIFO}
    \end{axis}
    \end{tikzpicture}
    \caption{Runtimes per sample for different FJSP sizes.}
    \label{fig:runtimes}
\end{figure}
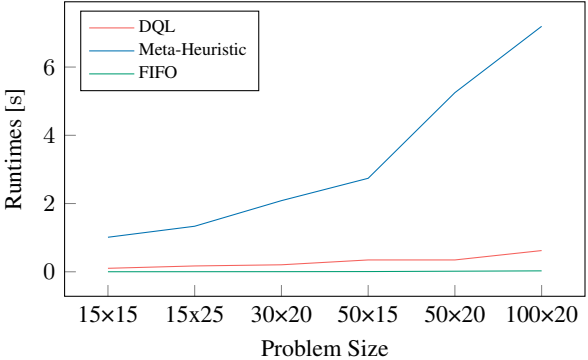

In \Cref{fig:runtimes}, we compare the runtime performance for different sizes of FJSP. While the meta-heuristic seems to increase in polynomial time, the runtime of DQL is nearly constant with problem size.

\section{Conclusion}\label{ccl}

In this work, we demonstrate how graph neural networks can be used to efficiently solve large, complex combinatorial optimization problems. By framing the CO instances as graph sequences, we can use reinforcement learning to find promising solutions. These solutions will be approximate, but while keeping the relative error low, we find our method scales much better in runtime than a more accurate meta-heuristic. We believe our results show promising initial results towards approaching the performance of state-of-the-art heuristic-based solvers.

\begin{ack}
DISTRIBUTION STATEMENT A. Approved for public release. Distribution is unlimited.
This material is based upon work supported by the Under Secretary of Defense for Research and Engineering under Air Force Contract No. FA8702-15-D-0001. Any opinions, findings, conclusions or recommendations expressed in this material are those of the author(s) and do not necessarily reflect the views of the Under Secretary of Defense for Research and Engineering.
© 2022 Massachusetts Institute of Technology.
Delivered to the U.S. Government with Unlimited Rights, as defined in DFARS Part 252.227-7013 or 7014 (Feb 2014). Notwithstanding any copyright notice, U.S. Government rights in this work are defined by DFARS 252.227-7013 or DFARS 252.227-7014 as detailed above. Use of this work other than as specifically authorized by the U.S. Government may violate any copyrights that exist in this work.
\end{ack}

\printbibliography

\end{document}